\documentclass[a4paper, fleqn]{cas-dc}

\usepackage[numbers]{natbib}
\usepackage{amsmath}
\usepackage{amssymb}
\usepackage{enumitem}
\usepackage{booktabs}

\def\tsc#1{\csdef{#1}{\textsc{\lowercase{#1}}\xspace}}
\tsc{WGM}
\tsc{QE}

\begin{document}
\let\WriteBookmarks\relax
\def\floatpagepagefraction{1}
\def\textpagefraction{.001}

\shorttitle{EANet: Expert Attention Network for Online Trajectory Prediction}    

\shortauthors{Pengfei Yao et al.}

\title[mode = title]{EANet: Expert Attention Network for Online Trajectory Prediction}

\tnotemark[1,2]

\tnotetext[1]{This document is the results of the research project funded by the National Science Foundation.}

\author[1]{Pengfei Yao}[type=editor,
    style=chinese,
    orcid=0009-0006-5526-8364]
\cormark[1] 
\fnmark[1] 
\ead{yaopengfei22@mails.ucas.ac.cn}
\author[1,4]{Tianlu Mao}[style=chinese]
\cormark[1]
\author[2]{Min Shi}[style=chinese]
\author[1,3]{Jingkai Sun}[style=chinese]
\author[1,4]{Zhaoqi Wang}[style=chinese]

\address[1]{Institute of Computing Technology Chinese Academy of Sciences, Beijing 100080, China}
\address[2]{North China Electric Power University,Beijing 100096, China}
\address[3]{Beijing University of Posts and Telecommunications, Beijing 100876, China}
\address[4]{Beijing Key Laboratory of Mobile Computing and
Pervasive Device, Beijing 100080, China}

\cortext[1]{Corresponding author}

\begin{abstract}
Trajectory prediction plays a crucial role in autonomous driving. Existing mainstream research and continuoual learning-based methods all require training on complete datasets, leading to poor prediction accuracy when sudden changes in scenarios occur and failing to promptly respond and update the model. Whether these methods can make a prediction in real-time and use data instances to update the model immediately(i.e., online learning settings) remains a question. The problem of gradient explosion or vanishing caused by data instance streams also needs to be addressed. Inspired by Hedge Propagation algorithm, we propose Expert Attention Network, a complete online learning framework for trajectory prediction. We introduce expert attention, which adjusts the weights of different depths of network layers, avoiding the model updated slowly due to gradient problem and enabling fast learning of new scenario's knowledge to restore prediction accuracy. Furthermore, we propose a short-term motion trend kernel function which is sensitive to scenario change, allowing the model to respond quickly. To the best of our knowledge, this work is the first attempt to address the online learning problem in trajectory prediction. The experimental results indicate that traditional methods suffer from gradient problems and that our method can quickly reduce prediction errors and reach the state-of-the-art prediction accuracy.
\end{abstract}


\begin{keywords}
Trajectory Prediction \sep 
Online Learning \sep 
Graph Neural Network
\end{keywords}
\maketitle

\section{Introduction}
The goal of trajectory prediction is to forecast a probable future trajectory using observed environmental information and the agents' motion trajectory, i.e. the coordinate sequence sampled at a previous fixed time interval\cite{sgan}. Accurate prediction of trajectory is critical in several applications including automated driving\cite{autodriving1, autodriving2}, intelligent robot decision planning\cite{tpnet, trajectron++}.

\begin{figure}[!ht]
  \centering
  \resizebox{0.4\textwidth}{!}{
    \includegraphics[scale=1]{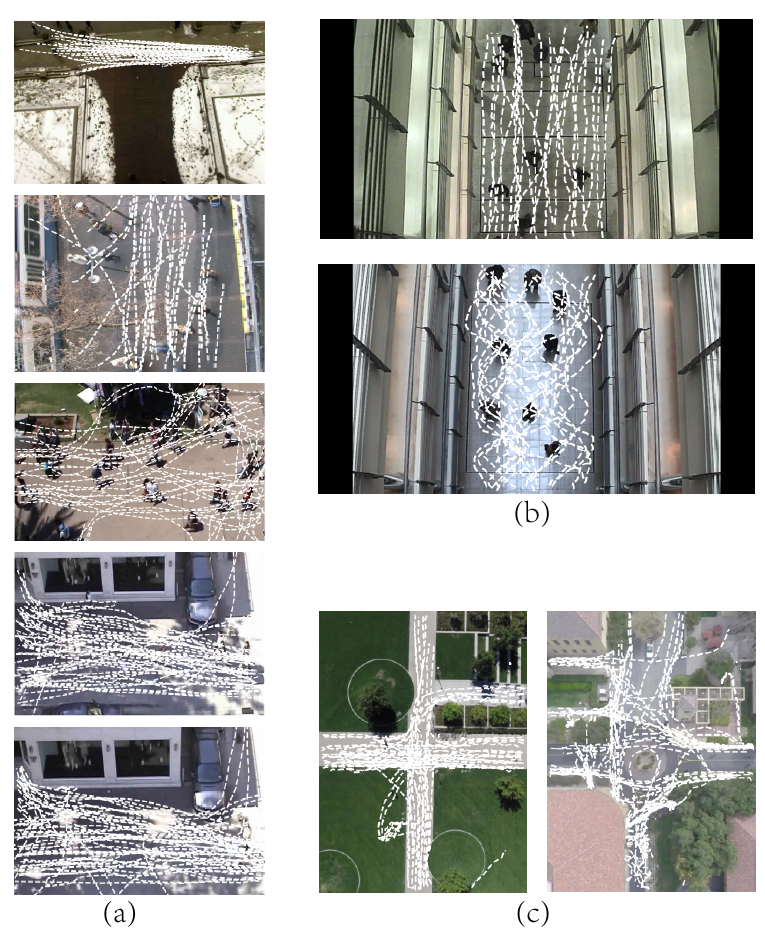}
  }
  \caption{(a) Visualization of the ETH-UCY, from top to bottom are eth, hotel, univ, zara1 and zara2. (b) Agents suddenly turn into a staggering walk when agents mainly went straight. (c) The sidewalk scene changes to a circle intersection scene.
  }
  \label{example}
\end{figure}

Currently, most existing trajectory prediction methods\cite{unlimited, ynet, personalized, stgcnn, sgcn}  
train models on one specific dataset to predict agent trajectory. 
Models trained in this manner are unable to maintain their original prediction accuracy when faced with changes in the environment\cite{habibi2020sila}.
Existing works on trajectory prediction based on continual learning are capable of achieving good prediction accuracy in new environments, which focus on addressing the method of continuous arrival of datasets in the form of sequences and how to prevent catastrophic forgetting of prior knowledge in the model. However, these researches are still based on offline training with a complete dataset.
To the best of our knowledge, there is currently no research focused on the online response of models to changes in the environment.

As depicted in Fig\ref{example}, when the scenario changes, or the agent's motion pattern and intention change, the model trained in the initial scenario will experience a varying degree of reduction in prediction accuracy\cite{habibi2020sila}. While data improvement strategies like trajectory rotation\cite{slstm, sgan} can be used to solve this problem, this is not always feasible in dynamic open environments, where scenario may differ significantly from training scenarios and may change over time. Such scenario changes, e.g. the changes in obstacles after a car accident at an intersection or the changes in pedestrian intentions after a fire in a public place, are common and sudden in dynamic open environments. 

However, updating the model based on newly arrived data instances often results in gradient explosion due to significant differences in data distributions, leading to problems of abnormal training and prediction of the model.
(This problem is described in detail in \textbf{Section A} of the appendix). 
And the existing continual learning methods still require sufficient data to be collected and complete training to be conducted for new environments. Furthermore, additional memory units are often used to address catastrophic forgetting, which increases the spatial overhead and computational complexity, making it difficult to respond to sudden changes in the scenario in real-time and quickly reduce prediction errors.

Online learning is a machine learning method based on data streams, in which each individual data instance is considered as the minimum unit of the data stream, as opposed to continual learning where the dataset is considered as the minimum unit. In online learning, the model is updated with every incoming data instance, while in continual learning, the model is updated after all the data instances are collected and formed into a dataset\cite{habibi2020sila}. This results in faster response to changes in the environment for trajectory prediction. Currently, most online learning methods are focused on classification tasks\cite{onlinesurvey}, but this does not imply that these methods can be directly applied to high-dimensional regression problems such as trajectory prediction. To the best of our knowledge, there are few online learning methods used for trajectory prediction in the field of autonomous driving.

In this paper, we propose Expert Attention Network (EANet), which use Expert Attention(EA) to enable Graph Convolutional Network (GCN) and Convolutional Neural Network (CNN) to have the capability of online learning. We design a kernel function for simulating the short-term trajectory motion trend to enhance the model's sensitivity to trajectory changes. Our method outperforms existing trajectory prediction methods in terms of adapting to scenario changes and is able to improve the prediction accuracy in new scenarios more rapidly compared to those based on continuous learning, with almost no additional resource consumption.

To summarize, our contributions are three folds:
\begin{itemize}[leftmargin=0.5cm, itemindent=0cm]
    \item[$\bullet$] To the best of our knowledge, our work is the first to investigate the online learning problems in agent trajectory prediction. We have experimentally validated that existing methods are unable to perform online learning in new scenarios. And we present a  online learning framework and evaluate its performance on a mutated scenario after being trained on a benchmark dataset.
    \item[$\bullet$] We propose an Expert Attention(EA) mechanism which adjusts the weights of shallow and deep layers in the network to control the output results and the parameter learning speed during back-propagation. EA enables the model to quickly learn when the scenario changes.
    \item[$\bullet$] We design a kernel function that can not only simulate the short-term trajectory motion trend, but also imitate the interactions between agents, and be sensitive to trajectory changes, in order to improve the efficiency of online learning.
\end{itemize}

\begin{figure*}[h!]
  \centering
  \includegraphics[width=0.95\linewidth]{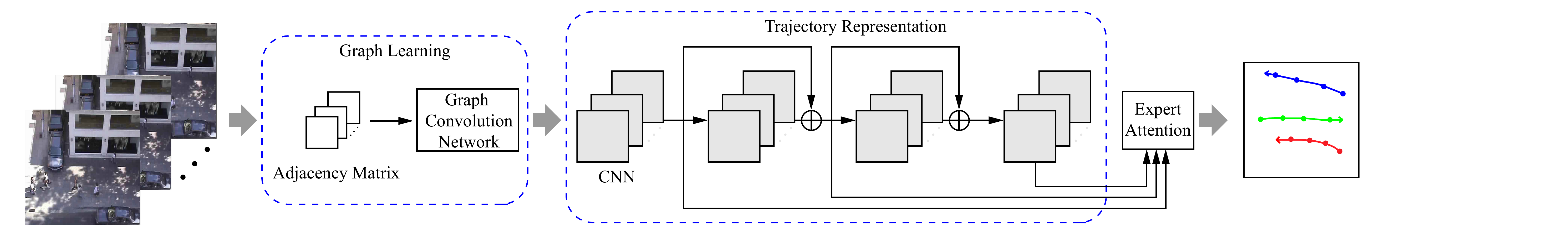}
  \caption{ The overall architecture of our proposed EANet.In the experiment, the number of stacked CNNs is set to 5.}
  \label{pipeline}
\end{figure*}

\section{Related Work}
\subsection{Trajectory Prediction}
Deep learning has become popular in trajectory prediction in recent years. Social-LSTM\cite{slstm} is one of the earlier works attempting to apply a deep learning model for trajectory prediction. Social-LSTM models pedestrian motion with Long Short-Term Memory\cite{lstm} (LSTM) and proposes a social pooling layer to calculate pedestrian interaction. Social-GAN\cite{sgan}, which introduces the  Generative Adversarial Network (GAN)\cite{gan}, proposes a new pooling mechanism to aggregate the human interaction information and solves the trajectory multimodality problem by predicting several trajectories and picking the best one. Sophie\cite{sophie} and Social Attention\cite{sattention} introduce the attention mechanism to assign different importance to different agents. 
Y-net\cite{ynet} structures trajectory predictions across long prediction horizons by modeling the epistemic and aleatoric uncertainty.

Since the introduction of the Graph Convolutional Network (GCN)\cite{gcn}, which is better suited to sparse node information in the scene, GCN has become a popular method for trajectory prediction. Graph convolution operation can quickly extract the features between nodes by weighted aggregation of the information of target nodes and adjacent nodes. 
STGAT\cite{stgat} builds spatial-temporal graphs and applies the GAT to model the interaction features. 
Social-STGCNN\cite{stgcnn}, which is the pipeline of our proposed EANet, generates a spatio-temporal graph from the agent trajectories in the scene and calculates the weighted adjacency matrix using kernel function to model the agent interaction. SGCN\cite{sgcn} proposes a sparse graph convolution network for pedestrian trajectory prediction, which uses the self-attention mechanism to generate a sparse representation of the graph. The majority of the work, however, is still focused on solving agent interactions and multimodal trajectory prediction challenges. When scenario changes, prediction accuracy suffers significantly, and due to batch training's inherent characteristics, the model is unable to effectively regain accuracy in the data flow.

\subsection{Continual Learning}
Continual learning focuses on developing methods for training models that can effectively learn from new data without forgetting previous knowledge. These approaches aim to overcome the challenge of catastrophic forgetting, which refers to the phenomenon where models forget previously learned knowledge when they are updated with new information.Some of the common strategies used in continual learning include rehearsal methods\cite{rehearsal}, regularization\cite{claw} and architecture\cite{architecture}. 

Some researches have explored the performance of trajectory prediction in continual learning. SILA\cite{habibi2020sila} is the first method for pedestrian trajectory prediction that utilizes incremental learning to adapt to changes in the scenario and maintain prediction accuracy. CLTP-MAN\cite{cltpman} uses memory augmented networks to increase the accuracy of predictions and adapt to changing scenes. Social-GR\cite{gr} presents a new approach in trajectory prediction by augmenting memory with a social generative replay mechanism. Hengbo Ma\cite{ma2021continual} utilizes a conditional generative memory model to handle the challenges of non-stationary data distributions. Luzia Knoedler\cite{knoedler2022improving} presents a self-supervised continual learning approach for improving pedestrian prediction models in dynamic environments.
Despite the fact that continual learning-based trajectory prediction methods can address the problem of catastrophic forgetting in new environments, they still require a large amount of data collection and complete offline training to be implemented in new environments. They are unable to respond to sudden changes in real-world scenarios. Furthermore, these methods often require additional storage of prior knowledge and datasets in the network, making it difficult to update the model in real-time.

\subsection{Online Learning}
Online learning\cite{onlinesurvey} is an important machine learning method. The learning strategy of online learning\cite{datastream} is to update the model parameters while predicting when the data instance arrives one by one. Online learning is more concerned with the model's real-time training effect, which is different from offline batch learning. Solving the phenomenon of concept drift is also an important research problem in online learning\cite{hedge}. OSAM\cite{osam} proposes a new online semi-supervised learning model, which has produced good classification results in the data flow as the amount of data and data categories have increased over time.
The hedge back-propagation algorithm\cite{hedge} designs a set of weights to calculate the gradient of each layer network based on the loss function of each layer's output and the real results, resulting in the capacity to resist concept drift. Online Bayesian Inference algorithm for CTR model\cite{ctr} proposes a novel inference method for learning from data streams. However, this kind of method often needs heuristic means to adjust parameters, and because these are abstract algorithms based on the deep model, it cannot be well executed when the scene and data distribution change vary significantly. In conclusion, to the best of our knowledge, there is no online learning method for trajectory prediction.

\section{Our Method}
Given the video frame $t=1,2,3\dots$, the two-dimensional spatial coordinates $(x^{i}_{t}, y^{i}_{t})$ of agents in each video frame, $i \in \lbrace1,2,\dots,N_{t}\rbrace$, where $N_{t}$ is the number of agents in the current video frame. The goal of the trajectory prediction is to predict the coordinate position sequence $\lbrace(x^{i}_{t},y^{i}_{t})|t=T_{obs}+1,T_{obs}+2,\dots,T_{obs}+T_{pred}\rbrace$ of agent $i$ for a future time period by observing the coordinate position sequence$\lbrace(x^{i}_{t},y^{i}_{t})|t=1,2,\dots,T_{obs}\rbrace$.

To solve the problem of prediction accuracy reduction caused by scenario change, we propose a short-term motion trend graph convolution network based on expert attention, which is categorized into three stages: graph learning, trajectory representation, and expert attention. Figure \ref{pipeline} shows the overall architecture of our proposed network. Firstly, to modal the extraction of motion trend features by graph convolution network, we propose a kernel function to calculate the influence of the short-term motion trend of trajectory. The calculation results are organized into a weighted adjacency matrix based on the graph constructed by the agent in current frame. This matrix adjusts the weights of edges in graph, making the trajectory feature extracted by GCN more accurate.  Then, the trajectory features are fed into CNNs for trajectory representation, and the output of each CNN layer is preserved and fed into expert attention.

\subsection{Graph Learning}

When the time is t, the graph formed of agent information is $G_{t}=(V_{t},E_{t},R_{t})$. $V_{t}=\lbrace v^{i}_{t}| \forall i\in\lbrace1, 2,\dots, N_{t}\rbrace\rbrace $ is the vertex set of $G_{t}$. Each agent is regarded as a vertex, which stores its coordinate $(x^{i}_{t},y^{i}_{t})$ and other information. $E_{t}=\lbrace e^{ij}_{t}|\forall i,j\in\lbrace1, 2,\dots, N_{t}\rbrace\rbrace$ is  the edge set of $G_{t}$. $e^{ij}_{t}$ exists if there is a relation between vertex $v^{i}_{t}$ and $v^{j}_{t}$. $R_{t}=\lbrace r^{i}_{t}|\forall i\in\lbrace1, 2, \dots, N_{t}\rbrace\rbrace$ is the motion trend set, where $r^{i}_{t}$ is the relative displacement between $v^{i}_{t-1}$ and $v^{i}_{t}$. 

\begin{figure*}[h!]
  \centering
  \includegraphics[width=0.95\linewidth]{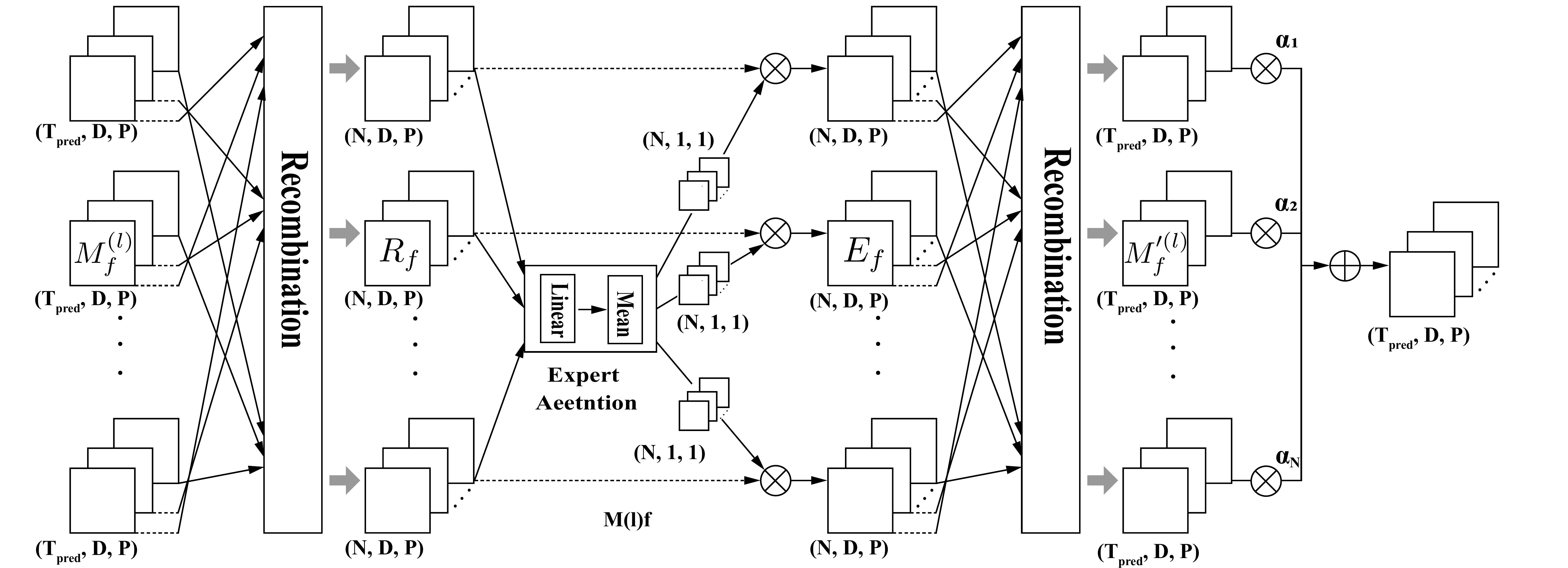}
  \caption{The implementation of expert attention. The expert attention in the figure is a shared module. Each $(N, D, P)$ will be calculated with the linear layer in the $D$ dimension, and then averaged in the $P$ dimension. In the experiment, the parameter N is set to 1, D is set to 5\cite{stgcnn}.}
  \label{expertattention}
\end{figure*}

The existing kernel functions\cite{stgcnn} mostly use the node distance to indicate the magnitude of the interaction between nodes. However, the agent motion inertance also plays an important role in agent interaction in the short-term motion. In fact, the motion state of the previous frame actually heavily influences the motion state of the following frame because the agent motion has inertia. The short-term interaction will be significantly impacted by the varying motion states of the various agents. We propose a kernel function, described in equation \ref{eq1}, to express the short-term trajectory motion trend and surrounding interaction of $v^{i}_{t}$ and $v^{j}_{t}$.
\begin{equation}
    \alpha^{ij}_{t}=\begin{cases}
\frac{1}{\left\|r^{i}_{t}-r^{j}_{t}\right\|_{2}+\left\|v^{i}_{t}-v^{j}_{t}\right\|_{2}}, &\left\|v^{i}_{t}-v^{j}_{t}\right\|_{2}\neq 0\\
0, &Otherwise \\
\end{cases}
\label{eq1}
\end{equation}
where $\left\|r^{i}_{t}-r^{j}_{t}\right\|_{2}$ denotes the short-term motion trend between two vertices. The higher influence of two vertices on each other, the stronger short-term motion trend is; $\left\|v^{i}_{t}-v^{j}_{t}\right\|_{2}$ represents the relative distance between two vertices. The interaction between two vertices will be more evident when the relative distance is closer. Our proposed kernel function can both model the interaction and the motion trends by composing the weighted adjacency matrix $A_{t}$. Especially, since $\left\|r^{i}_{t}-r^{j}_{t}\right\|_{2}$ simply takes into account the motion trend from the previous frame, it can indicate the differences in motion states of various agents at various moments, which means $\alpha^{ij}_{t}$ is sensitive to trajectory change.

We symmetrically normalize the $A_{t}$ obtained by the kernel function according to equation \ref{eq2} to ensure GCN work properly\cite{stgcnn}.
\begin{equation}
A_{t}^{'}=D^{-1/2}(A_{t}+I)D^{-1/2}
\label{eq2}
\end{equation}
where $I$ is the identity matrix of vertices and $D$ is the graph degree matrix. The graph convolution can therefore be written as equation \ref{eq3}.
\begin{equation}
g(V_{t}, A_{t}) = \sigma(D^{-1/2}(A_{t}+I)D^{-1/2}V_{t}\textbf{W})
\label{eq3}
\end{equation}
where $\textbf{W}$ is the trainable parameter in GCN, $\sigma$ is sigmoid function. By using the graph convolution that utilizes the normalized weighted adjacency matrix of the short-term motion trend, we can effectively extract the short-term motion features of the trajectory in space, and provide a more accurate feature for trajectory representation.

\subsection{Trajectory Representation}

Different from graph learning which mainly describes spatial features, trajectory representation expresses temporal features. We define the feature tensor yielded by GCN with size $(N,T_{obs},P,D)$, where $N$ is the batch size of the input network, $T_{obs}$ is the number of observed trajectory frames, $P$ is the number of agents in the current scene, and $D$ is the prediction dimension. In trajectory representation, to begin, a layer of CNN is used to shift the number of frames from $T_{obs}$ to $T_{pred}$, which is the number of predicted trajectory frames. Then we stack CNNs in order to extract more detailed spatio-temporal feature information and increase prediction accuracy.

\subsection{Expert Attention}

\begin{table*}[]
\centering
\resizebox{\textwidth}{!}{%
\begin{tabular}{cccccccccccc}
\hline
\multirow{2}{*}{Model} &
  \multirow{2}{*}{Years} &
  \multirow{2}{*}{ETH-UCY} &
  \multicolumn{3}{c}{SNU} &
  \multicolumn{6}{c}{SDD} \\ \cmidrule(r){4-6} \cmidrule(r){7-12} 
 &
   &
   &
  Oneway &
  Stroll &
  Stagger &
  bookstore &
  coupa &
  deathCircle &
  gates &
  hyang &
  nexus \\ \hline
S-LSTM &
    2016 &
    0.72/1.54 &
    0.49/1.02 &
    1.01/2.05 &
    0.65/1.32 &
    0.72/1.41 &
    0.47/0.93 &
    0.79/1.58 &
    0.68/1.38 &
    0.51/1.07 &
    0.71/1.45 \\
Trajectron++  &
    2020 &
    0.53/1.11 &
    0.26/0.53 &
    0.83/1.66 &
    0.43/0.85 &
    0.52/1.05 &
    0.35/0.69 &
    0.61/1.19 &
    0.46/0.96 &
    0.34/0.68 &
    0.52/1.03 \\
Social-STGCNN &
    2020 &
    0.54/1.12 &
    0.27/0.53 &
    0.83/1.68 &
    0.43/0.85 &
    \underline{0.51}/\underline{1.03} &
    0.34/0.67 &
    0.63/1.24 &
    0.48/1.00 &
    0.36/0.69 &
    0.50/0.98 \\
EqMotion &
  2023 &
  \underline{0.49}/\underline{1.03} &
  \underline{0.25}/\underline{0.48} &
  \underline{0.78}/\underline{1.51} &
  \textbf{0.40}/\textbf{0.79} &
  0.52/1.03 &
  \underline{0.30}/\underline{0.58} &
  \underline{0.57}/\underline{1.15} &
  \textbf{0.43}/\textbf{0.87} &
  \underline{0.35}/\underline{0.71} &
  \underline{0.47}/\underline{0.90} \\
EANet(Ours) &
  / &
  \underline{0.50}/\textbf{1.01} &
  \textbf{0.24}/\textbf{0.47} &
  \textbf{0.76}/\textbf{1.49} &
  \underline{0.41}/\underline{0.83} &
  \textbf{0.50}/\textbf{1.02} &
  \textbf{0.29}/\textbf{0.57} &
  \textbf{0.55}/\textbf{1.12} &
  \underline{0.45}/\underline{0.88} &
  \textbf{0.33}/\textbf{0.67} &
  \textbf{0.46}/\textbf{0.87} \\ \hline
\end{tabular}%
}
\caption{Comparison with the different methods on different datasets for ADE/FDE. The lower the better. The best result is in bold. The second best result is underlined.
}
\label{base_result_table}
\end{table*}

We increase model depth by stacking CNNs in trajectory representation so that the model could mine deeper features. However, in online learning, a ResNet-like\cite{resnet} architecture will become a limitation: The shallow network can quickly update the parameters to convergence, but the prediction accuracy will be inadequate. The deep network can achieve higher model prediction accuracy through more training epochs, but the performance of it is inferior to the shallow network when there are few data instances or when the training period is short. What's worse, in online learning which train model by only one data instance, deep model will result in unstable updating: When the scenario changes, the model is vulnerable to damage from gradient exploding or gradient vanishing.

To overcome the challenge of stacked networks, we developed expert attention, which uses expert weight to make full use of outputs from different layers, enabling model to converge quickly without being damaged by gradient problem and maintain the ability to increase the prediction accuracy. When scenario changes, expert attention makes model more inclined to the shallow network, ensuring prediction accuracy quickly returning to normal; after a period of scenario change, the expert attention will tend toward the deep layer, which can extract more depth feature information, guaranteeing that prediction accuracy does not suffer.

Fig\ref{expertattention} shows the implementation of expert attention. We begin by gathering the results of each intermediate outputs in trajectory representation, which has a size of $(N, T_{pred}, P, D)$. Then we recombine those outputs indexing by time frame. We assume that the the number of CNN in model is $L$. $M^{(l)}_{f}$ stands for the track of the $f$-th frame output by the $l$-layer network, the recombination is:
\begin{equation}
R_f=concat(\lbrace M^{(l)}_{f}|l=1, 2, \dots, L\rbrace)
\label{eq4}
\end{equation}
where $f=1, 2, \dots, T_{pred}$, $R_{f}$ is the module containing all intermediate outputs' $f$ frame results, which has a size of $(N, L, P, D)$. According to equation \ref{eq4}, we calculate the expert weight.
\begin{equation}
E_{f}=Mean(\tanh{(R_{f}\Theta^{T}+b)})
\label{eq5}
\end{equation}
where $E_{f}$ is the expert weight score, which has the size $(N, L, 1, 1)$. $Mean(*)$ is the pedestrian-indexed mean function. $\Theta^{T}$ is a $(D, 1)$ dimensional matrix. After computing each pedestrian's score via full connection, the average value is determined on the pedestrian dimension to acquire the attention distribution of each layer's output results. The expert weight is then multiplied with the recombination module by extending it to the same size as the recombination module.
\begin{equation}
M'^{(l)}_{f} = concat(\lbrace R^{(l)}_{f}\odot E^{'}_{f}|f=1, 2, \dots, T_{pred}\rbrace ) 
\label{eq6}
\end{equation}
where $l=1, 2, \dots, L$, $M'^{(l}_{f}$ is the trajectory representation of weighted recombined to the original size, $E^{'}_{f}$ is an extension of $E_{f}$ in dimension. Finally, using a one-dimensional weighted vector, we reversely merge the results of expert attention back to the original intermediate output form and obtain the final trajectory output by multiplying it with a one-dimensional weighted vector with the same layer size $(\alpha_{1}, \alpha_{2}, \dots, \alpha_{L})$, which is trainable.
\begin{equation}
Out = \sum_{l = 1}^{L}\alpha_{l}M'^{(l)}
\label{eq7}
\end{equation}
where $Out$ is the final output of the model, which is the probability estimation of sequence coordinates. By weighting different frames with shared expert attention and weighting different layers with simple weight vector, the model can quickly adjust the weight distribution according to the current trajectory prediction.

\section{Experiment and Analysis}

\subsection{Experimental Settings}
Our experimental environment is: Ubuntu 20.04, Python3.8, Pytorch 1.11.0, CUDA 11.5, NVIDIA GeForce RTX 3060. In our experiment, Hyperparameter settings of offline training for baseline model refer to the original paper, making the comparison of experiment valuable. As for EAnet, we set the number of stacking layers in trajectory representation to 5. The model is trained for 250 epochs using Stochastic Gradient Descent (SGD) with a batch size of 128. The learning rate starts at 0.01 and drops to 0.002 after 150 epochs. To control the variables in experiment, we offline train all models on benchmark dataset ETH-UCY. In Online Learning experiment, we avoid the learning failure described in Appendix \textbf{A} by using some prior techniques, such as gradient clipping and scenario alignment, enabling comparative experiment to be conducted. The training batch size in online learning is 1 to simulate data flow. We train 1000 data instances in each scene. 

\subsection{Loss Function and Evaluation Metrics}
\label{metrics}
We assume that the trajectory coordinates prediction follows a binary Gaussian distribution. For the trajectory coordinates $(x^{i}_{t}, y^{i}_{t})$ of agent $i$ at time $t$, the model's ultimate output is the trajectory coordinate distribution $\mathcal{N}(\tilde{\mu}^{i}_{t} , \tilde{\sigma}^{i}_{t}, \tilde{\rho}^{i}_{t})$ of agent I at time t, where $\tilde{\mu}^{i}_{t}$ is the mean, $\tilde{\sigma}^{i}_{t}$ is the standard deviation, and $\tilde{\rho}^{i}_{t}$ is the correlation coefficient. The negative log-likelihood loss function is minimized to train the model, which is defined as:
\begin{equation}
L^{i}(\textbf{W})=-\sum_{t=1}^{T_{pred}}log(\mathbb{P}((x^{i}_{t}, y^{i}_{t})|\tilde{\mu}^{i}_{t} , \tilde{\sigma}^{i}_{t}, \tilde{\rho}^{i}_{t}))
\label{eq8}
\end{equation}

We utilize two metrics to measure the model's prediction accuracy: Average Displacement Error (ADE) and Final Displacement Error (FDE)\cite{slstm}. These two metrics can accurately measure the model's capacity. In this experiment, we use 8 frames of historical trajectory to predict 12 frames. More details can be found in the appendix B.
\begin{equation}
    ADE=\frac{1}{N\times T_{pred}}\sum_{i=1}^{N}\sum_{t=1}^{T_{pred}}\left\|\tilde{v}^{i}_{t}-v^{i}_{t}\right\|_{2}, 
    \label{eq9}
\end{equation}
\begin{equation}
    FDE=\frac{1}{N}\sum_{i=1}^{N}\left\|\tilde{v}^{i}_{T_{pred}}-v^{i}_{T_{pred}}\right\|_{2}
    \label{eq10}
\end{equation}

In addition to the commonly ADE/FDE indicators in trajectory prediction, we design another indicator to evaluate the effectiveness of model's capacity of online learning: restore\_ratio(rr),  the result is the proportion of disparity between the prediction error of the model predicting after $n$ instances’ online learning and the prediction error of the model offline training and predicting in that scenario.
\begin{equation}
    rr_{n} = \frac{1}{2}(\frac{ADE_{n} - ADE_{base}}{ADE_{base}} + \frac{FDE_{n} - FDE_{base}}{FDE_{base}})
    \label{restore_ratio}
\end{equation}
where $ADE_{base}$ is the ADE of model directly offline trained on new scenario. $n=0$ is model directly predicting on new scenario after offline trained on ETH-UCY. 
\subsection{Evaluation Datasets}
\label{dataset}

We employed three trajectory datasets, ETH-UCY\cite{ucy, eth}, SNU\cite{snu}, and SDD\cite{sdd}. ETH-UCY is the most extensively used benchmark dataset in pedestrian trajectory prediction. SNU focuses on the indoor scene and delivers pedestrian trajectory data in a variety of patterns, including one-way, stagger, stroll, etc. SDD is a large-scale dataset with a variety of agents (pedestrians, cyclists, skateboards, etc.) and 29 real-world outdoor scenarios, such as a school campus. We imitate the process of scenario change using batch training on ETH-UCY and online learning on SNU or SDD to meet our research need. We use the average results of the five scenarios of ETH-UCY and the average results of each sub-scenario of SDD to evaluate.

\subsection{Online Experimental Framework}
\begin{table*}[]
\centering
\resizebox{\textwidth}{!}{%
\begin{tabular}{ccccccccccc}
\hline
\multicolumn{2}{c}{\multirow{2}{*}{Model}} &
  \multicolumn{3}{c}{ETH-UCY$\rightarrow$SNU} &
  \multicolumn{6}{c}{ETH-UCY$\rightarrow$SDD} \\ \cmidrule(r){3-5} \cmidrule(r){6-11}  
\multicolumn{2}{c}{} &
  $\rightarrow$Oneway &
  $\rightarrow$Stroll &
  $\rightarrow$Stagger &
  $\rightarrow$bookstore &
  $\rightarrow$coupa &
  $\rightarrow$deathCircle &
  $\rightarrow$gates &
  $\rightarrow$hyang &
  $\rightarrow$nexus \\ \hline
\multirow{5}{*}{Begining} &
  S-LSTM &
  0.87/1.75 &
  1.44/2.91 &
  0.96/1.99 &
  1.21/2.47 &
  0.83/1.68 &
  1.18/2.40 &
  1.06/2.16 &
  0.91/1.82 &
  1.13/2.17 \\
 &
  Trajectron++ &
  0.60/1.22 &
  1.22/2.43 &
  0.76/\underline{1.52} &
  1.07/2.20 &
  0.62/1.28 &
  0.97/1.91 &
  0.82/1.66 &
  0.78/1.60 &
  0.99/1.99 \\
 &
  Social-STGCNN &
  0.63/1.28 &
  1.23/\underline{2.39} &
  0.77/1.58 &
  \underline{1.01}/\underline{1.98} &
  0.57/\underline{1.20} &
  0.96/1.87 &
  0.79/1.58 &
  0.76/1.54 &
  0.96/1.94 \\
 &
  EqMotion &
  \underline{0.56}/\underline{1.10} &
  \underline{1.20}/2.43 &
  \textbf{0.74}/\textbf{1.50} &
  1.05/2.09 &
  \underline{0.56}/1.21 &
  \underline{0.92}/\underline{1.83} &
  \textbf{0.77}/\underline{1.52} &
  \underline{0.74}/\underline{1.46} &
  \underline{0.94}/\textbf{1.81} \\
 &
  EANet(Ours) &
  \textbf{0.55}/\textbf{1.09} &
  \textbf{1.17}/\textbf{2.32} &
  \underline{0.75}/1.53 &
  \textbf{0.99}/\textbf{1.94} &
  \textbf{0.52}/\textbf{1.07} &
  \textbf{0.91}/\textbf{1.81} &
  \textbf{0.77}/\textbf{1.51} &
  \textbf{0.71}/\textbf{1.41} &
  \textbf{0.93}/\underline{1.82} \\ \hline
\multirow{5}{*}{Online trained} &
  S-LSTM &
  0.72/1.39 &
  1.28/2.55 &
  0.80/1.65 &
  1.04/2.19 &
  0.66/1.32 &
  1.05/2.04 &
  0.93/1.80 &
  0.75/1.46 &
  0.99/1.83 \\
 &
  Trajectron++ &
  0.41/0.80 &
  1.05/2.04 &
  0.60/1.14 &
  0.91/1.80 &
  0.45/0.94 &
  0.77/1.53 &
  0.63/1.26 &
  0.59/1.26 &
  0.81/1.63 \\
 &
  Social-STGCNN &
  0.41/0.88 &
  1.01/\underline{2.01} &
  \underline{0.55}/\underline{1.12} &
  \underline{0.81}/\underline{1.58} &
  \underline{0.36}/\underline{0.70} &
  0.74/1.45 &
  0.60/1.16 &
  0.58/\underline{1.10} &
  \underline{0.74}/1.54 \\
 &
  EqMotion &
  \underline{0.40}/\underline{0.68} &
  \underline{1.00}/\underline{2.01} &
  0.58/1.13 &
  0.86/1.73 &
  0.39/0.79 &
  \underline{0.72}/\underline{1.43} &
  \underline{0.58}/\underline{1.12} &
  \underline{0.56}/1.12 &
  \underline{0.74}/\underline{1.45} \\
 &
  EANet(Ours) &
  \textbf{0.22}/\textbf{0.43} &
  \textbf{0.82}/\textbf{1.66} &
  \textbf{0.42}/\textbf{0.87} &
  \textbf{0.57}/\textbf{1.12} &
  \textbf{0.20}/\textbf{0.37} &
  \textbf{0.58}/\textbf{1.13} &
  \textbf{0.42}/\textbf{0.85} &
  \textbf{0.38}/\textbf{0.71} &
  \textbf{0.49}/\textbf{0.98} \\ \hline
\end{tabular}
}
\caption{Quantitative comparison of models using ADE/FDE. The lower the better. The best result is in bold. The second best result is underlined. \textit{begining} rows indicate the prediction error of each model after offline training on ETH-UCY and directly switching to other scenarios. \textit{Online trained} rows indicate the prediction error of each model after 1000 instances arrived for online learning.}
\label{begining and after}
\end{table*}
The two parts of our procedure are offline training and online learning.  As the EANet, first, we offline train the graph learning and trajectory representation on the ETH-UCY. Then in online learning, the expert attention is initialized according to the number of trajectory representation layers. As other baseline, we offline train the complete model. We online train the model on SNU or SDD with batch size 1 to simulate data flow. We choose Social-LSTM\cite{slstm}, Social-STGCNN\cite{stgcnn}, Trajectron++\cite{trajectron++} and EqMotion\cite{eqmotion} for comparison. We comprehensively compared the results of direct training and prediction of these models on various scenarios, and the results of online learning on SDD and SNU scenarios after offline training on ETH-UCY. What's more, We also compared the effects of different kernel functions on model prediction and the results of direct application of existing online learning strategies\cite{hedge}, which described in Appendix C, on trajectory prediction problems to more comprehensively illustrate the advantages of our method.

\begin{figure}[!ht]
  \centering
  \resizebox{0.45\textwidth}{!}{
    \includegraphics[scale=1]{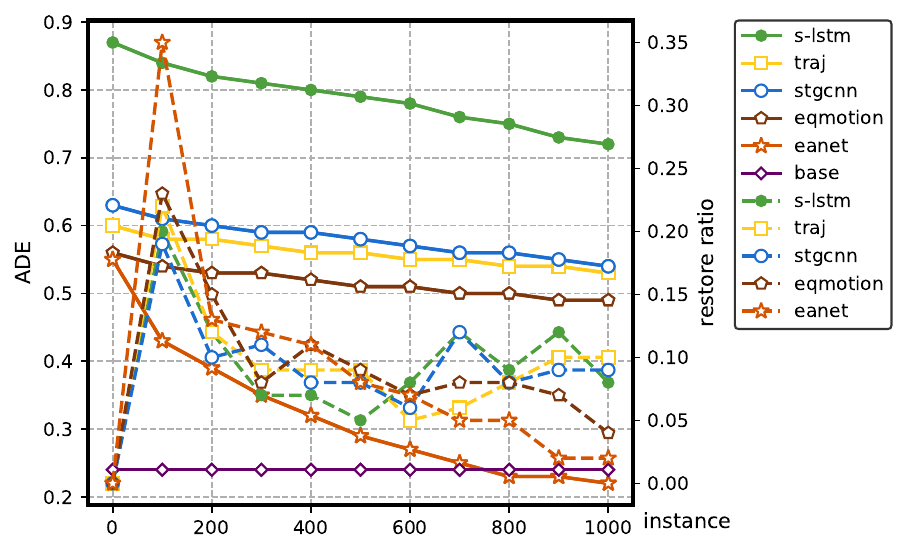}
  }
  \caption{
  The ADE/FDE and $rr$ curves of different methods when online learning on Oneway. The solid line represents the line of ADE/FDE, and the dashed line represents the line of $rr$. base represents the best model directly offline training and predicting on Oneway.
  }
  \label{rr_fig}
\end{figure}

\subsection{Quantitative Analysis}

\begin{table*}[]
\centering
\resizebox{0.8\textwidth}{!}{
\begin{tabular}{cccc}
\hline
Model &
  $\rightarrow$Oneway &
  $\rightarrow$Stroll &
  $\rightarrow$Stagger \\ \hline
S-LSTM &
   74.56\% $\rightarrow$ 67.97\% $\rightarrow$ 41.61\% &
   42.26\% $\rightarrow$ 38.92\% $\rightarrow$ 25.56\% &
   49.22\% $\rightarrow$ 44.18\% $\rightarrow$ 24.04\% \\
Trajectron++   &
   130.48\% $\rightarrow$ 117.25\% $\rightarrow$ 54.32\% &
   46.69\% $\rightarrow$ 42.29\% $\rightarrow$ 24.70\% &
   77.78\% $\rightarrow$ 70.59\% $\rightarrow$ 36.83\% \\
Social-STGCNN &
   137.42\% $\rightarrow$ 120.72\% $\rightarrow$ 58.94\% &
   45.23\% $\rightarrow$ 38.32\% $\rightarrow$ 20.66\% &
   82.48\% $\rightarrow$ 70.59\% $\rightarrow$ 29.84\% \\
EqMotion &
   126.58\% $\rightarrow$ 111.43\% $\rightarrow$ 50.83\% &
   57.39\% $\rightarrow$ 52.04\% $\rightarrow$ 30.66\% &
   87.44\% $\rightarrow$ 76.76\% $\rightarrow$ 44.02\% \\
EANet(Ours) &
   \textbf{130.54\%} $\rightarrow$ \textbf{83.90\%} $\rightarrow$\textbf{-8.42}\% &
   \textbf{54.83\%} $\rightarrow$ \textbf{50.05\%} $\rightarrow$ \textbf{9.65\%} &
   \textbf{83.63\%} $\rightarrow$ \textbf{54.63\%} $\rightarrow$ \textbf{3.63\%} \\ \hline
Model &
  $\rightarrow$bookstore &
  $\rightarrow$coupa &
  $\rightarrow$deathCircle \\ \hline
S-LSTM &
   71.62\% $\rightarrow$ 66.27\% $\rightarrow$ 49.88\% &
   78.62\% $\rightarrow$ 72.13\% $\rightarrow$ 41.18\% &
   50.63\% $\rightarrow$ 46.71\% $\rightarrow$ 31.01\% \\
Trajectron++ &
   107.65\% $\rightarrow$ 102.76\% $\rightarrow$ 73.21\% &
   81.33\% $\rightarrow$ 69.55\% $\rightarrow$ 32.40\% &
   59.76\% $\rightarrow$ 51.29\% $\rightarrow$ 27.40\% \\
Social-STGCNN &
   95.14\% $\rightarrow$ 88.33\% $\rightarrow$ 56.11\% &
   73.38\% $\rightarrow$ 58.74\% $\rightarrow$ 5.18\% &
   51.59\% $\rightarrow$ 46.71\% $\rightarrow$ 17.20\% \\
EqMotion &
   102.42\% $\rightarrow$ 96.27\% $\rightarrow$ 66.67\% &
   97.64\% $\rightarrow$ 84.73\% $\rightarrow$ 33.10\% &
   60.27\% $\rightarrow$ 53.28\% $\rightarrow$ 25.33\% \\
EANet(Ours) &
   \textbf{94.10\%} $\rightarrow$ \textbf{66.33\%} $\rightarrow$ \textbf{11.90\%} &
   \textbf{83.51\%} $\rightarrow$ \textbf{40.71\%} $\rightarrow$\textbf{-33.06}\% &
   \textbf{63.53\%} $\rightarrow$ \textbf{40.41\%} $\rightarrow$ \textbf{3.17\%} \\ \hline
Model &
  $\rightarrow$gates &
  $\rightarrow$hyang &
  $\rightarrow$nexus \\ \hline
S-LSTM &
   56.20\% $\rightarrow$ 52.68\% $\rightarrow$ 33.60\% &
   74.26\% $\rightarrow$ 69.76\% $\rightarrow$ 41.75\% &
   54.41\% $\rightarrow$ 50.09\% $\rightarrow$ 32.82\% \\
Trajectron++   &
   75.59\% $\rightarrow$ 67.30\% $\rightarrow$ 34.10\% &
   132.35\% $\rightarrow$ 121.76\% $\rightarrow$ 79.41\% &
   91.79\% $\rightarrow$ 84.83\% $\rightarrow$ 57.01\% \\
Social-STGCNN &
   61.29\% $\rightarrow$ 52.13\% $\rightarrow$ 20.50\% &
   117.15\% $\rightarrow$ 107.77\% $\rightarrow$ 60.27\% &
   94.98\% $\rightarrow$ 86.40\% $\rightarrow$ 52.57\% \\
EqMotion &
   76.89\% $\rightarrow$ 66.87\% $\rightarrow$ 31.81\% &
   108.53\% $\rightarrow$ 96.60\% $\rightarrow$ 58.87\% &
   100.56\% $\rightarrow$ 92.21\% $\rightarrow$ 59.28\% \\
EANet(Ours) &
   \textbf{71.35\%} $\rightarrow$ \textbf{46.61\%} $\rightarrow$\textbf{-5.04}\% &
   \textbf{112.80\%} $\rightarrow$ \textbf{78.02\%} $\rightarrow$ \textbf{10.56\%} &
   \textbf{105.68\%} $\rightarrow$ \textbf{71.94\%} $\rightarrow$ \textbf{9.58\%} \\ \hline
\end{tabular}%
}
\caption{Quantitative comparison of methods using $rr$, which described in section\ref{metrics}. In this table, we provide $rr_{0}\rightarrow rr_{100}\rightarrow rr_{1000}$. The best result group is in bold.}
\label{restore}
\end{table*}

Table \ref{base_result_table} shows the results of each benchmark model's direct training and prediction on each scenario, showing that each model are trained well. the model trained on ETH-UCY will serve as the pre-trained model for upcoming online learning experiment, and the results of  others will use for restore\_ratio comparison. Moreover, Table\ref{base_result_table} shows that our proposed method obtains the best or suboptimal results on all datasets, which indicates that EANet can accurately predict trajectories, and the prediction error is better than most methods.

Quantitative comparison of ADE/FDE in online learning and restore\_ratio are respectivly shown on Table \ref{begining and after} and Table \ref{restore}. The \textit{beginning} part of Table \ref{begining and after} shows that our proposed methods can still achieve the best prediction arcuracy in most scenarios when directly predicting in new scenarios. The \textit{Online Learning} part of Table \ref{begining and after} indicates that EANet obtains the best prediction accuracy in all scenarios after online learning in few instances, and the prediction accuracy is far better than other methods. Quantitatively,  EANet improves ADE/FDE 27.19\% on SNU and 31.35\% on SDD after learning in 1000 instances. 

Table \ref{restore} futher illustrates the advantages of EANet in recovering prediction accuracy. EANet can quickly reduce the prediction error of the model while retaining the ability of further learning, and even make the model achieve a lower prediction error than the base result, which can be seen from results of Oneway and gates, the $rr$ can reach negative after online learning.

The ADE/FDE and $rr$ curves of different methods when online learning on Oneway are shown in Fig \ref{rr_fig}. We can see EANet is able to learn new scenarios far faster than other models, and eventually can further improve the prediction accuracy of deep models. Overall, the prediction accuracy restore speed of EANet 191.67\% faster than existing methods.

\section{Computational and Memory Consumption}
\begin{table}[!ht]
    \setlength{\tabcolsep}{9pt}
    \centering
    \begin{tabular}{ccc}
    \toprule
    Method        & Memory/KB & Computation/fps  \\
    \midrule
    Trajectron++  &           &                  \\
    Social-STGCNN & 46.2      & 42.5             \\
    EqMotion      & 24166.4   &                  \\
    EANet(Ours)   & 46.8      & 29.4             \\
    \bottomrule
    \end{tabular}
    \caption{Computational and Memory Consumption.}
    \label{consumpiton}
\end{table}
As shown in Table \ref{consumpiton}, we statistically evaluate the computational and memory consumption of using our method. We notice that compared with Social-STGCNN\cite{stgcnn}, which has good speed and accuracy in batch training, our method consumes only 1.3\% additional memory. As for computational consumption, although our attention will add a lot of computation, the prediction rate of the method can still reach 30 frames per second, which can meet the speed requirements of online real-time prediction.

\subsection{Ablation Study}
\begin{table}[]
\centering
\resizebox{\columnwidth}{!}
{
    \begin{tabular}{cccccc}
        \hline
        \multicolumn{4}{c}{Kernel} &
        \multicolumn{2}{c}{$\rightarrow$ SDD} \\ \cmidrule(r){1-4} \cmidrule(r){5-6} 
        $\beta^{ij}_{t}$ &  $\gamma^{ij}_{t}$ & $\epsilon^{ij}_{t}$ &   $\alpha^{ij}_{t}$ & ADE/FDE &   rr \\ \hline
        \checkmark       &                    &                     &                     &0.55/1.12|1.03/2.05$\rightarrow$0.78/1.54    &85.15\%$\rightarrow$79.33\%$\rightarrow$39.66\% \\
        &\checkmark &   &   &  0.59/1.20|1.13/2.25$\rightarrow$0.81/1.63 &89.51\%$\rightarrow$80.57\%$\rightarrow$36.56\% \\
        &  &\checkmark &    &0.61/1.21|1.14/2.29$\rightarrow$0.87/1.71 &88.07\%$\rightarrow$80.16\%$\rightarrow$41.97\% \\
        &  &  & \checkmark & \textbf{0.43}/\textbf{0.85}|\textbf{0.81}/\textbf{1.60}$\rightarrow$ \textbf{0.44}/\textbf{0.86} &  \textbf{88.30}\%$\rightarrow$ \textbf{58.06}\%$\rightarrow$ \textbf{1.75}\%  \\ \hline
    \end{tabular}
}
\caption{Ablation study of model using different kernel function. In \textit{ADE/FDE} column, the left side of the vertical line shows the results of offline training and predicting on ETH-UCY, the other side shows the prediction error from directly predicting on other scenario to predicting after 1000 instances' online learning, using the model offline trained on ETH-UCY. The meaning of \textit{rr} column is described in section \ref{metrics}. The best result group is in bold.
}
\label{ablation1}
\end{table}

\begin{table}[]
\centering
\resizebox{\columnwidth}{!}
{
    \begin{tabular}{cccc}
    \hline
    \multicolumn{2}{c}{Online Strategy} & \multicolumn{2}{c}{$\rightarrow$ SNU}                                                               \\ \cmidrule(r){1-2} \cmidrule(r){3-4} 
    Hedge    & EA   & ADE/FDE                                     & restore\_ratio                                         \\ \hline
    &   &   0.47/0.93|0.76/1.52$\rightarrow$ 0.67/1.33 &  64.7\%$\rightarrow$ 61.41\%$\rightarrow$ 42.78\% \\
    \checkmark &  & \multicolumn{1}{l}{0.47/0.93|0.93/1.84$\rightarrow$ 0.68/1.34} &  97.86\%$\rightarrow$ 95.07\%$\rightarrow$ 65.89\% \\
             & \checkmark                        & \textbf{0.47/0.93}|\textbf{0.82/1.65}$\rightarrow$ \textbf{0.49/0.99} &  \textbf{75.94}\%$\rightarrow$ \textbf{63.36}\%$\rightarrow$ \textbf{5.35}\%  \\ \hline
    \multicolumn{2}{c}{Online Strategy} & \multicolumn{2}{c}{$\rightarrow$ SDD}                                                                                 \\ \cmidrule(r){1-2} \cmidrule(r){3-4}
    Hedge    & EA   & ADE/FDE                                     & restore\_ratio                                         \\ \hline
    &   &   0.43/0.85|0.74/1.49 $\rightarrow$ 0.64/1.28  &   73.69\% $\rightarrow$ 70.1\% $\rightarrow$ 49.71\%   \\
    \checkmark        &                          & 0.43/0.85|0.88/1.72$\rightarrow$ 0.64/1.27 &  103.5\%$\rightarrow$ 96.35\%$\rightarrow$ 49.12\% \\
             & \checkmark                        & \textbf{0.43}/\textbf{0.85}|\textbf{0.81}/\textbf{1.60}$\rightarrow$ \textbf{0.44}/\textbf{0.86} &  \textbf{88.30}\%$\rightarrow$ \textbf{58.06}\%$\rightarrow$ \textbf{1.75}\%  \\ \hline
    \end{tabular}
}
\caption{Ablation study of different online strategy. The meanings of each data are the same as those in Table \ref{ablation1}. The best result group is in bold.}
\label{ablation2}
\end{table}

Tables \ref{ablation1} and \ref{ablation2} provide the smile experiment results for kernel functions and online learning strategies, respectively. In Table \ref{ablation1}, we compare the distance reciprocal kernel function $\beta^{ij}_{t}$\cite{stgcnn}, $L2$ norm $\gamma^{ij}_{t}$ and Gaussian Radial Basis Function $\epsilon^{ij}_{t}$\cite{gaussiankernel} to our proposed kernel function. Table \ref{ablation1} shows that our proposed kernel function quickly restore prediction accuracy by incorporating the short-term motion trend. Table \ref{ablation2} indicates that, compared to directly training model in new scenario, Hedge back-propagation algorithm has no significant effect in online learning, and our proposed EA can quickly reduce the prediction error while further improving the prediction capacity of the model.
\begin{equation}
\beta^{ij}_{t}=\begin{cases}
\frac{1}{\left\|v^{i}_{t}-v^{j}_{t}\right\|_{2}}, &\left\|v^{i}_{t}-v^{j}_{t}\right\|_{2}\neq 0\\
0, &Otherwise \\
\end{cases}   
\label{eq13}
\end{equation}
\begin{equation}
\gamma^{ij}_{t}=\begin{cases}
\left\|v^{i}_{t}-v^{j}_{t}\right\|_{2}, &\left\|v^{i}_{t}-v^{j}_{t}\right\|_{2}\neq 0\\
0, &Otherwise \\
\end{cases}   
\label{eq11}
\end{equation}
\begin{equation}
\epsilon^{ij}_{t}=\begin{cases}
\frac{exp(-\left\|v^{i}_{t}-v^{j}_{t}\right\|_{2})}{\sigma}, &\left\|v^{i}_{t}-v^{j}_{t}\right\|_{2}\neq 0\\
0, &Otherwise \\
\end{cases}   
\label{eq12}
\end{equation}

\subsection{Qualitative Analysis}
\begin{figure}[!ht]
  \centering
  \resizebox{0.35\textwidth}{!}{
    \includegraphics[scale=1]{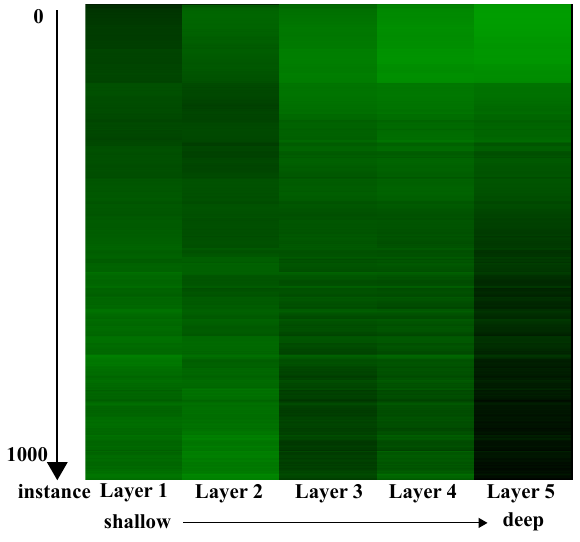}
  }
  \caption{
  Visualization of expert attention weights shift between shallow and deep layer connected with scenario change. The darker the green, the greater the weight. 
  }
  \label{ea}
\end{figure}

Figure \ref{ea} illustrates that when scenario changes, the attention will be oriented toward the shallow layer. After the deep model’s prediction accuracy is recovered and improved, the attention will become smooth and close to the deep layer, resulting in the effect of combating scenario change.

Trajectory visualization result of different methods in online learning can be found in Appendix D.

\section{Conclusion}
\label{conclusion}

In this paper, we propose an expert attention network for online trajectory prediction in dynamic open environment. Our proposed method can maintain good performance in the situation of significant changes in scenario. According to thorough experimental evaluation, our method can swiftly restore the model's prediction accuracy and converge to a lower error after scenario changed. The proposed kernel function models the short-term motion trend of agents and the relative position between agents, allowing the model to generate more consistent trajectory features. Our proposed expert attention balances the output of the shallow and deep networks to manage the speed of parameter update, preventing gradient exploding when scenario changes while maintaining the accuracy advantage of a deep model in data flow training. However, the kernel function is limited to graph networks, and expert attention is only focused on networks stacked of the same size. It does not have a universal applicability in methods based on LSTM or RNN.

\appendix
\section{Online Learning in Traditional Methods}
We chose representative trajectory prediction methods with different base network architectures (LSTM, CNN, GCN), including Social-LSTM\cite{slstm}, STGAT\cite{stgat}, Y-net\cite{ynet}, and Social-STGCNN\cite{stgcnn}. These models were trained offline on ETH-UCY\cite{eth, ucy} and subjected to online learning on SDD\cite{sdd} across various scenarios. Specifically, each time a data instance arrived, the model updated its predictions and itself. Table \ref{online base result} shows the problem when directly online train the traditional method. From the results, it can be concluded that the existing methods cannot be directly applied to cross-scenario online learning.

\begin{table*}[!ht]
\setlength{\tabcolsep}{9pt}
\label{online base result}
\centering
\resizebox{\textwidth}{!}{
    \begin{tabular}{lccccccc}
\toprule
\multirow{2}{*}{Model} & \multicolumn{6}{c}{SDD}                                                                                       \\     \cmidrule(r){2-7} 
                       & bookstore:4     & coupa:1       & deathCircle:5  & gates:8           & hyang:4           & nexus:7            \\ \midrule
Social-LSTM            & 9/184/7(4.5\%)  & 5/6/39(10\%)  & 0/250/0(0\%)   & 11/42/347(2.8\%)  & 35/68/97(17.5\%)  & 18/161/171(5.1\%)  \\ 
STGAT                  & 16/173/11(8\%)  & 7/2/41(14\%)  & 6/244/0(2.4\%) & 16/39/345(4\%)    & 38/51/111(19\%)   & 29/134/187(8.3\%)  \\ 
Ynet                   & 13/180/7(6.5\%) & 6/11/31(12\%) & 7/243/0(2.8\%) & 23/58/319(5.8\%)  & 51/64/85(25.5\%)  & 34/188/128(9.7\%)  \\ 
Social-STGCNN          & 22/169/9(11\%)  & 6/4/40(12\%)  & 9/241/0(3.6\%) & 27/42/331(6.75\%) & 49/49/102(24.5\%) & 36/139/175(10.3\%) \\ \bottomrule
\end{tabular}
}
\caption{Online learning results of the methods on various scenarios of the SDD dataset. The results are based on 50 experiments per scenario to alleviate randomness. The numbers following each scenario indicate the number of sub-scenarios; the numbers separated by slashes in the results denote the occurrences of three cases: "the model was able to learn and converge normally", "the gradient exploded during training, causing the model to fail to converge or even crash", and "the gradient vanished during training, making the model unable to update". The numbers in parentheses represent the ratio of the cases where the model was able to learn online normally.}
\end{table*}

\section{Evaluation Metrics}
We utilize two metrics to measure the model's prediction accuracy for the output results of the training model: Average Displacement Error(ADE) and Final Displacement Error(FDE). ADE refers to the average error between each agent's predicted temporal position coordinates and all ground truth future trajectory points. FDE refers to the position coordinates of the last frame of each agent predicted by the model and the position coordinates of the true endpoint. These two metrics can accurately measure the model's trajectory prediction accuracy. Because the model's direct output is the probability distribution of trajectory coordinates, we construct numerous trajectories using probability distribution sampling and select the one that is closest to the ground truth for evaluation\cite{sgan, stgcnn}.

\begin{figure*}
  \centering
  \includegraphics[width=0.95\linewidth]{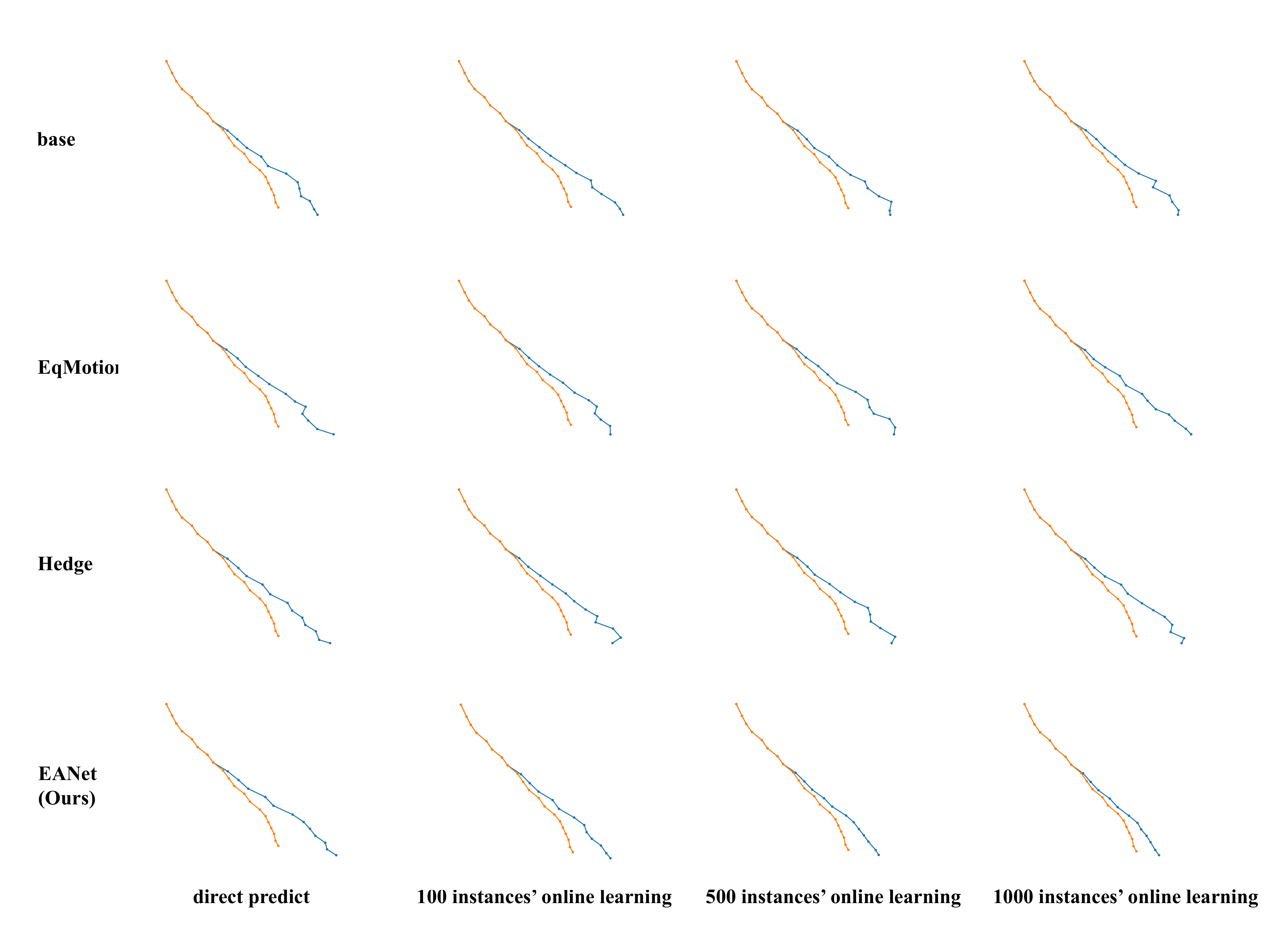}
  \caption{Visualization on gates\_6. Orange represents ground truth, and blue represents prediction.}
  \label{visual1}
\end{figure*}
\section{Hedge Back-propagation}
Hedge back-propagation algorithm is similar to Mixture of Expert approach in concept. On average, this procedure will set the expert weight. It collects the intermediate output of each layer of the stacked network during the prediction process and uses the weighted sum result as the model's final output. It adjusts the expert weight based on the loss function between the output result of each layer and the real result after obtaining the real data. As a consequence, the network can withstand concept drift. However, the parameters of this algorithm must be heuristically adjusted depending on the scene or data distribution, or it will have no discernible effect.

\section{Visualization}
Figure \ref{visual1} shows the trajectory visualization of different online strategy and models using online learning in gates\_6 scenario. We can see that our proposed method can quickly learn new scenario knowledge after 100 instances' online learning , approximate the real trajectory, and ultimately outperform the base case.

\bibliographystyle{cas-model2-names}

\bibliography{cas-refs}

\end{document}